\documentclass{article}

\usepackage{amsmath,amsfonts,bm}

\def\eqref#1{equation~\ref{#1}}

\def\1{\bm{1}}

\DeclareMathAlphabet{\mathsfit}{\encodingdefault}{\sfdefault}{m}{sl}
\SetMathAlphabet{\mathsfit}{bold}{\encodingdefault}{\sfdefault}{bx}{n}

\usepackage{microtype}
\usepackage{times}
\usepackage{epsfig}
\usepackage{subfigure}
\usepackage{graphicx}
\usepackage{amsmath}
\usepackage{amssymb}
\usepackage{setspace}
\usepackage{booktabs}       
\usepackage{makecell}

\usepackage{hyperref}

\usepackage[accepted]{sysml2019}

\sysmltitlerunning{Explaining Away Attacks Against Neural Networks}

\begin{document}

\twocolumn[
\sysmltitle{Explaining Away Attacks Against Neural Networks}




\begin{sysmlauthorlist}
\sysmlauthor{Sean Saito}{to}
\sysmlauthor{Jin Wang}{to}

\end{sysmlauthorlist}

\sysmlaffiliation{to}{SAP Asia, Singapore}

\sysmlcorrespondingauthor{Sean Saito}{sean.saito@sap.com}
\sysmlcorrespondingauthor{Jin Wang}{jin.wang02@sap.com}

\sysmlkeywords{Machine Learning, SysML}

\vskip 0.3in

\begin{abstract}
We investigate the problem of identifying adversarial attacks on image-based neural networks. We present intriguing experimental results showing significant discrepancies between the explanations generated for the predictions of a model on clean and adversarial data. Utilizing this intuition, we propose a framework which can identify whether a given input is adversarial based on the explanations given by the model. Code for our experiments can be found here: \url{https://github.com/seansaito/Explaining-Away-Attacks-Against-Neural-Networks}.
\end{abstract}
]

\printAffiliationsAndNotice{} 

\section{Introduction}
Adversarial attacks cause state-of-the-art neural networks to make misclassifications to inputs with imperceptible perturbations \cite{szegedy2013intriguing, goodfellow2014explaining}. In this work, we propose a framework for detecting adversarial attacks by utilizing the explanations of the model's predictions. Our experiments in Section \ref{sec:experiments} demonstrate that this framework can detect untargeted attacks at \textbf{99.81\%} accuracy and targeted attacks at \textbf{99.87\%} accuracy.

\begin{figure}[!ht]
\centering
\includegraphics[width=.8\linewidth]{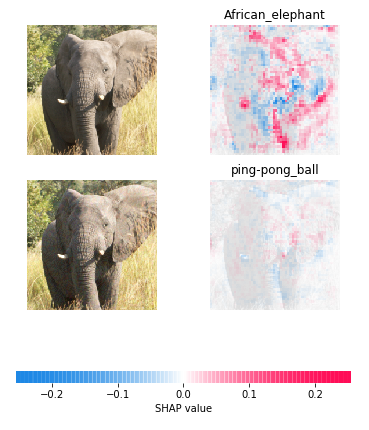}
\vspace{-5mm}
\caption{A comparison of SHAP Integrated Gradients computed for a pair of clean and adversarial images. The prediction for \textit{African\_elephant} had a confidence of \textbf{99.00\%}, whereas the prediction for \textit{ping-pong\_ball} was \textbf{99.21\%}.}

\label{fig:juxtaposition}
\end{figure}

\section{Explaining predictions on adversarial examples}
\label{sec:explaining_adv_examples}

In an adversarial attack, the goal of an attacker is to construct an input $\Tilde{x}$ that is minimally perturbed from the original clean input $x$ which causes some misclassification (an \textit{untargeted attack}) or a desired prediction (a \textit{targeted attack}) from a neural network. Several methods exist which achieve this, including the Basic Iterative Method (BIM) \cite{kurakin2016adversarial}, which iteratively finds the gradient of the model's loss function with respect to the input:

\vspace{-4mm}
\begin{equation*}
\begin{aligned}
    x_0 &= x \\
    \tilde{x_{i+1}} &= x_i - \epsilon \cdot sign(\nabla_{x_i} J(x_i, y_t)) \\
\end{aligned}
\end{equation*}
\vspace{-4mm}

where $\epsilon$ controls the magnitude of peturbations made to the image, resulting in perceptively identical images which cause large deviations in the behavior of the model. We hypothesize that the explanations of the model on each prediction would also vary significantly, which could potentially be a clue to identify whether a given prediction was caused by an adversarial attack.

To evaluate this hypothesis, we generate explanations via the SHAP Integrated Gradients framework \cite{NIPS2017_7062}, a framework which combines Integrated Gradients, a gradient-based pixel attribution method \cite{sundararajan2017axiomatic}, with SHAP, an additive feature attribution method that calculates Shapley values for each pixel:

\vspace{-4mm}
\begin{equation*}
\begin{aligned}
IG_i(x) &= \frac{x_i - b_i}{m} \sum_{h=1}^m
\frac{\partial f}{\partial x_i}( b + \frac{h}{m}(x-b) ) \\
\end{aligned}
\end{equation*}
\vspace{-4mm}

In all experiments in this work, we generate explanations via the Inception-V3 model \cite{szegedy2016rethinking}. Figure \ref{fig:juxtaposition} displays an example of explanations generated for a pair of clean and adversarial images. Red and blue values indicate the SHAP value of each pixel. The figure indicates qualitatively that the SHAP values for the clean image predictions are larger than those of the adversarial image.

\begin{figure}[!ht]
\centering
\includegraphics[width=1.0\linewidth]{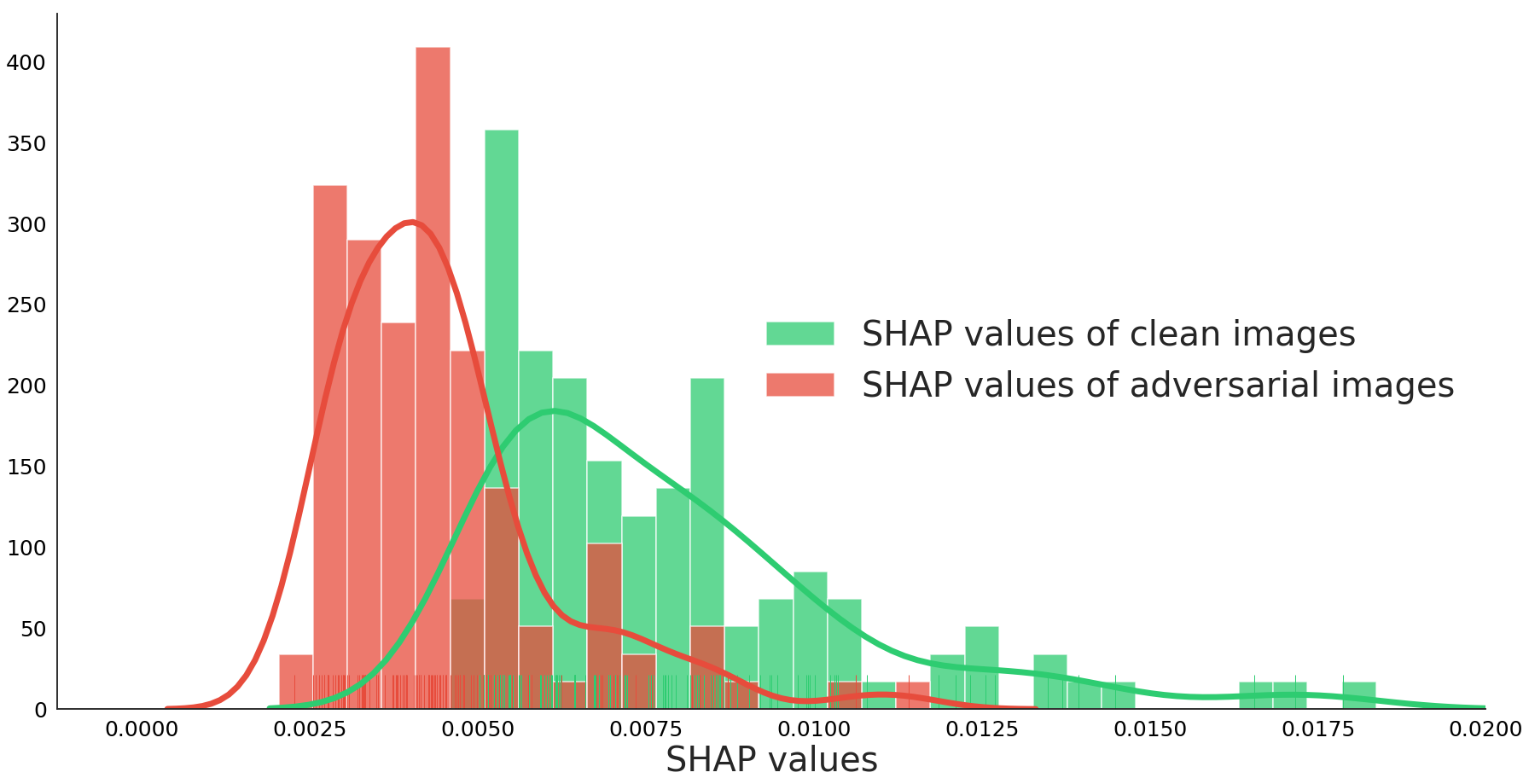}
\vspace{-5mm}
\caption{A comparison of SHAP Integrated Gradients distributions between clean images and adversarial images.}
\label{fig:distribution}
\end{figure} 

Figure \ref{fig:distribution} shows a distributional comparison of the absolute SHAP values across 1,000 test images from the ImageNet \cite{imagenet_cvpr09} dataset. We again observe a significant difference in the distribution between the SHAP values of clean and adversarial images. This suggests that neural networks are unable to provide sufficient "evidence" when explaining its decisions on adversarial examples. This intuition leads to our proposed framework of detecting adversarial attacks.

\section{A framework for detecting adversarial examples}
\label{sec:proposal_framework}

We propose a simple yet effective framework for detecting adversarial attacks. Based on the training images ($S_{ori}$), adversarial samples ($S_{adv}$), and neural network $f$, we treat the SHAP values coming from the predictions of $f$ as feature descriptors, denoted as $Z_{ori}$ and $Z_{adv}$ respectively. We combine $Z_{ori}$ and $Z_{adv}$ to construct a binary classification dataset which trains an adversarial example detector $D(z)$. For any future sample $s$ and prediction $f(s)$, we generate SHAP values $z_s$ and call $D(z_s)$ to indicate whether the sample is adversarial.

\section{Experiments and results}
\label{sec:experiments}
Our dataset contains 10,000 pairs of images from ImageNet across all classes, with 20\% held out for testing. We employ three types of BIM attacks: untargeted, targeted (random class), and targeted ($2^{nd}$ most confident class). We set $\epsilon$ as 0.1 for all attacks. Table \ref{table:success} shows attack success rates. 

We experiment with two off-the-shelf classifiers to assess how well they can detect adversarial examples based on the SHAP feature descriptors. Results shown in Table \ref{table:acc} indicate that simple classifiers can detect adversarial examples with high accuracy.

\begin{table}[!ht]
\centering
\caption{Attack success rate of each attack}
\vspace{1mm}
\scalebox{0.9}{
\begin{tabular}{lll}
\toprule
            & Success Rate\\
\midrule
Untargeted attack  &  99.13\%      \\
Targeted attack (random) & 98.41\% \\
Targeted attack ($2^{nd}$ most confident) & 99.87\% \\
\bottomrule
\end{tabular}
}
\label{table:success}
\end{table}

\begin{table}[!ht]
\centering
\caption{Accuracy of detecting each attack}
\vspace{1mm}
\scalebox{0.9}{
\begin{tabular}{llll}
\toprule
            &  \makecell{Logistic \\ Regression} & \makecell{Random \\ Forest} \\
\midrule
Untargeted attack  &  99.16\%   & 99.81\%  \\
Targeted attack (random) &98.96\%    & 99.83\% \\
Targeted attack ($2^{nd}$ confident) & 99.17\% & 99.87\% \\
\bottomrule
\end{tabular}
}
\label{table:acc}
\end{table}

\section{Conclusion and future work}
We have shown empirical results suggesting that explanations can help us detect adversarial attacks. Future work would focus on evaluating and refining this method under additional threat models.
\vspace{-4mm}

\bibliography{paper}
\bibliographystyle{sysml2019}


\end{document}